\documentclass[runningheads]{llncs}
\usepackage{graphicx}
\graphicspath{{./images/}}
\usepackage{comment}
\usepackage{amsmath,amssymb} 
\usepackage{epsfig}
\usepackage{subcaption}
\usepackage{lipsum}
\usepackage[dvipsnames]{xcolor}

\usepackage{booktabs}

\usepackage[breaklinks=true,bookmarks=false]{hyperref}

\begin{document}
\pagestyle{headings}
\mainmatter
\def\ECCVSubNumber{3640}  

\title{Visual Question Answering on Image Sets} 

\titlerunning{ISVQA}
%
\author{Ankan Bansal\inst{1}\thanks{This work was done when Ankan Bansal was an intern at AWS.} \and
Yuting Zhang\inst{2} \and
Rama Chellappa\inst{1}}
\authorrunning{A. Bansal et al.}
%
\institute{University of Maryland, College Park, \quad\email{\{ankan,rama\}@umd.edu} \and
Amazon Web Services (AWS), \quad\quad\email{yutingzh@amazon.com}
}
\maketitle

\begin{abstract}
We introduce the task of Image-Set Visual Question Answering (ISVQA), which generalizes the commonly studied single-image VQA problem to multi-image settings. Taking a natural language question and a set of images as input, it aims to answer the question based on the content of the images. The questions can be about objects and relationships in one or more images or about the entire scene depicted by the image set. To enable research in this new topic, we introduce two ISVQA datasets -- indoor and outdoor scenes. They simulate the real-world scenarios of indoor image collections and multiple car-mounted cameras, respectively. The indoor-scene dataset contains 91,479 human-annotated questions for 48,138 image sets, and the outdoor-scene dataset has 49,617 questions for 12,746 image sets. We analyze the properties of the two datasets, including question-and-answer distributions, types of questions, biases in dataset, and question-image dependencies. We also build new baseline
models to investigate new research challenges in ISVQA.  
\end{abstract}

\section{Introduction}
\label{sec:intro}


Answering natural-language questions about images requires understanding both linguistic and visual data. 
Since its introduction~\cite{Antol2015}, Visual Question Answering (VQA) has attracted
significant attention. 
Several related datasets~\cite{Marino2019,Zhu2016,Hudson2019} and methods~\cite{Gao2019,Lin2015,Goyal2019} have been proposed. 

\setcounter{footnote}{0}

\begin{figure}
    \begin{center}
        \includegraphics[width=0.45\linewidth]{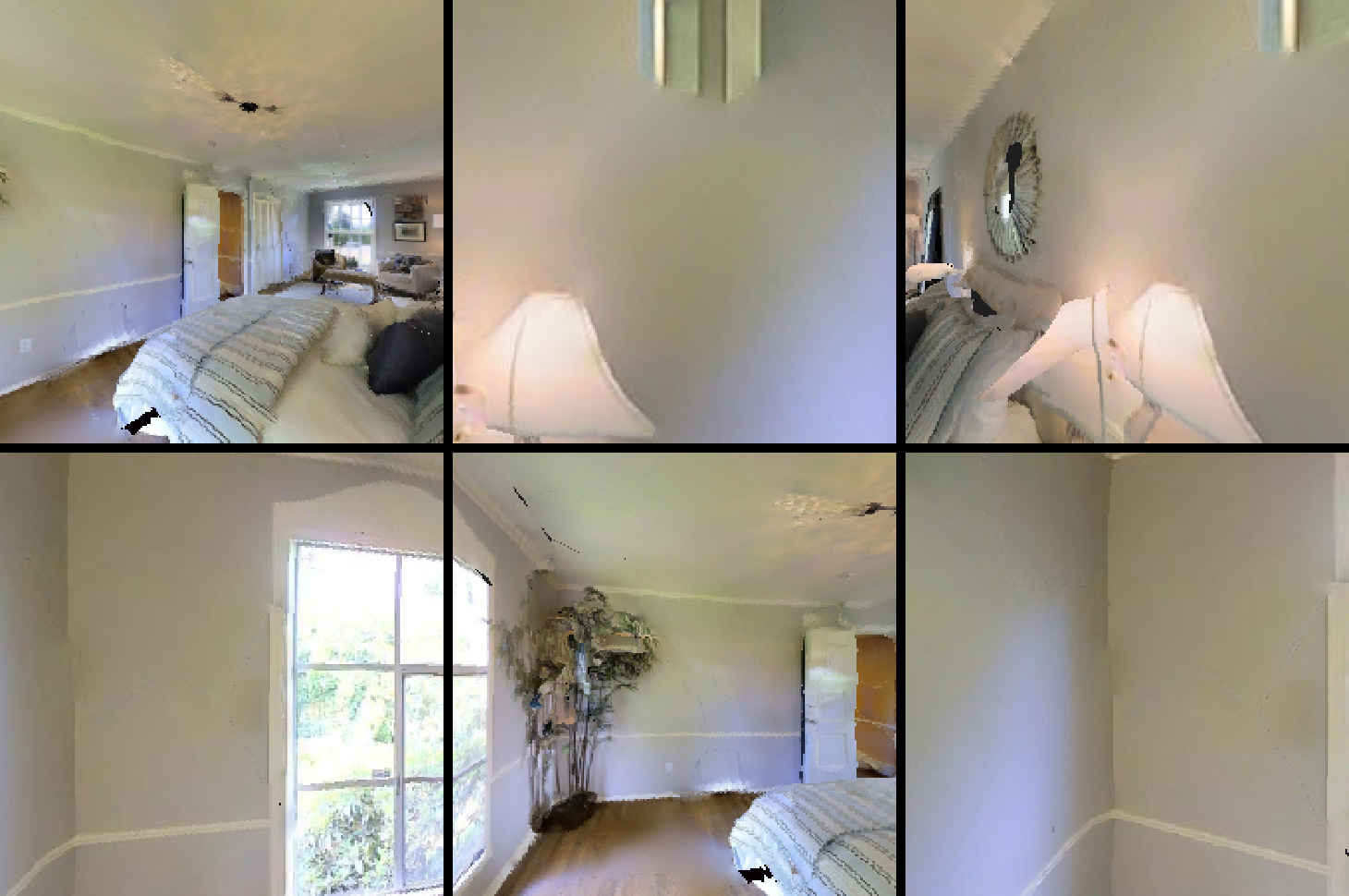}~~~~~~~
        \includegraphics[width=0.45\linewidth]{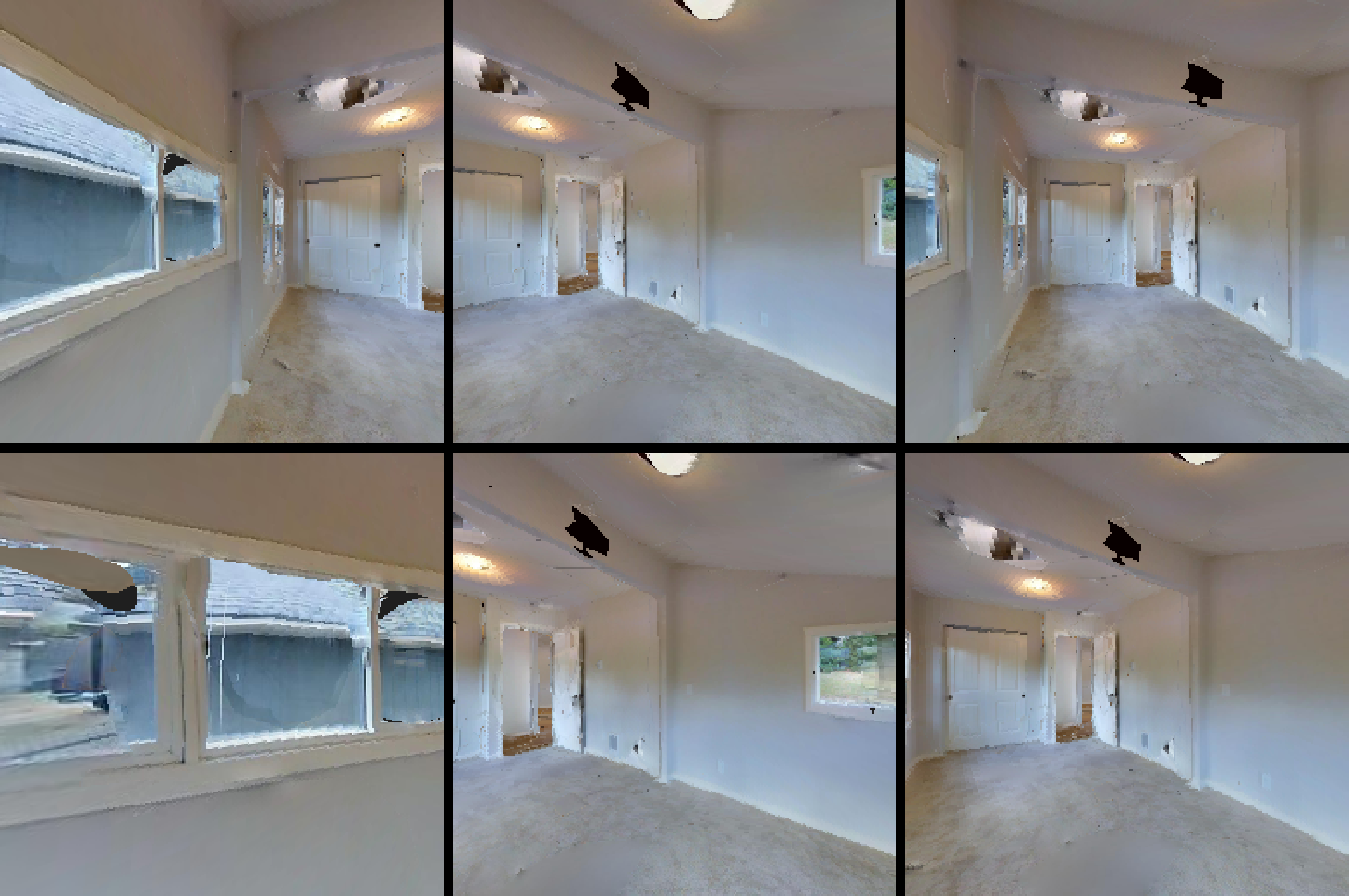}
    \end{center}
    \caption{(Left) Given the set of images above, and the question ``What is hanging above the bed?", it
    is necessary to connect the bed in the top-left image to the mirror in the top-right image. To
answer this question a model needs to understand the concepts of ``bed", ``mirror", ``above",
``hanging", etc. and be able to relate the bed in the first image with the headrest and pillows in
the third image. (Right) When asked the question ``How many rectangles are on the interior doors?", the model
    should be able to provide the answer (``four") and avoid counting the same rectangles multiple
times. }
    \label{fig:pull}
\end{figure}

In this paper, we introduce the new task of Image Set Visual Question Answering (ISVQA)
\footnote{Project page: \url{https://ankanbansal.com/isvqa.html}}. It aims to
answer a free-form natural-language question based on a set of images. The proposed ISVQA task
requires reasoning over objects and concepts in different images to predict the correct answer.
For example, for figure~\ref{fig:pull} (Left), a model has to find the relationship between the
\texttt{bed} in the top-left image and the \texttt{mirror} in the top-right, via \texttt{pillows}
which are common to both the images. This example shows the unique challenges associated with
image-set VQA. A model for solving this type of problems has to understand the question, find the
connections between the images, and use those connections to relate objects across images.
Similarly, in figure~\ref{fig:pull} (Right), the model has to avoid double-counting recurring objects in
multiple images.
These challenges associated with scene understanding have not been explored in existing single-image
VQA settings but frequently happen in the real world. 

ISVQA reflects information retrieval from multiple images of relevance but with no obvious
continuous correspondences. Such image sets can be any albums and images captured by multiple
devices, \emph{e.g.}, images under the same story on Facebook/Instagram, images of the same product
on Craigslist and Amazon, pictures of the same house on real estate websites, and images from
different car-mounted cameras. Other instances of the ISVQA task include answering questions about
images taken at different times (\emph{e.g.} like in camera trap photography), at different
locations (\emph{e.g.} multiple cameras from an indoor or outdoor location), or from different
viewpoints (\emph{e.g.} live sports coverage). Some of these settings contain images taken from the
same scene, while others might involve images of a larger span. 
While ISVQA can be generally applied to any type of images, in this paper, we focus on images from
multiple views of an environment, especially street and indoor scenes. 


ISVQA may require finding the same objects in different images or determining the
relationships between different objects within or across images. It can also entail determining
which images are the most relevant for the question and then answering based only on
them, ignoring the other images. 
ISVQA leads to new research challenges, including:  a) How
to use natural language to guide scene understanding across multiple views/images; and b) how to
fuse information from relevant images to reason about relationships among entities. 


To enable research into these problems, we built two datasets for ISVQA - one for indoor scenes and
the other for outdoor scenes. The indoor scenes dataset comes from Gibson Environment
\cite{xia2018gibson} and contains 91,479 human-generated questions, each for a set of images - for a
total of 48,138 image sets. Similarly, the outdoor scenes dataset comprises of 49,617 questions for
12,746 image sets. The images in the outdoor scenes dataset come from the nuScenes dataset
\cite{Caesar2019}. We explain the data collection methodology and statistics in section
\ref{sec:dataset}.


The indoor ISVQA dataset contains two parts: 1.) Gibson-Room - containing images from the
same room; and 2.) Gibson-Building - containing images from different places in the same building.
This is to facilitate spatial and semantic reasoning both in localized and extended regions
in the same scene. The outdoor dataset contains image sets taken from mostly urban
environments.

We propose two extensions of single-image VQA methods as baseline approaches to investigate the
ISVQA task and the datasets. In addition, we also use an existing Video VQA approach as a simple
baseline. Finally, we also propose to use use a transformer-based approach which can specifically
target ISVQA.
Such baselines meet significant difficulties in solving the ISVQA problem, and they reflect the
particular challenges of the ISVQA task.  We also present the statistics of the datasets, by
analyzing the types of question, distributions of answers for different types of questions, and
biases present in the dataset. 

In summary, we make the following major contributions in this work.
\begin{itemize}
    \item[-] We propose ISVQA as a new setting for scene understanding via Q\&A;
    \item[-] We introduce two large-scale datasets for targeting the ISVQA problem. In total, these
        datasets contain 141,096 questions for 60,884 sets of images. 
    \item[-] We establish baseline methods on ISVQA tasks to recognize the challenges and encourage
        future research. 
\end{itemize}


%
%
%

\section{Related Works}
\label{sec:related}



\noindent
\textbf{VQA settings}. The basic VQA setting 
\cite{Antol2015} involves answering natural language questions about images. The VisualGenome dataset
\cite{krishna2017visual} also contains annotations for visual question-answer pairs at both image
and region levels. Visual7W \cite{Zhu2016} built upon the basic VQA setting and introduced visual
grounding to VQA. Several other VQA settings target specific problems or applications. For example,
    VizWiz \cite{gurari2018vizwiz} was designed to help answer questions asked by blind people.
    RecipeQA \cite{yagcioglu2018recipeqa} is targeted for answering questions about recipes from
    multi-modal cues. TallyQA \cite{Acharya2019}, and HowMany-QA \cite{trott2018interpretable}
    specifically target counting questions for single images. Unlike these, the CLEVR
    \cite{johnson2017clevr} benchmark and dataset uses synthetically generated images of rendered 3D
    shapes and is aimed towards understanding the geometric relationships between objects. IQA
    \cite{Gordon2018} is also a synthetic setting where an agent is required to navigate a scene and
    reach the desired location in order to answer the question.

Unlike existing work, ISVQA targets scene understanding by answering questions which
might require multiple images to answer. This important setting has not been studied before and
necessitates a specialized dataset. Additionally, answering most of the questions requires a model
to ignore some of the images in the set. This capability is absent from many state-of-the-art VQA
models. 

We also distinguish our work from video VQA. Unlike many such datasets (\emph{e.g.}
        TVQA+ \cite{Lei2019}, MovieQA \cite{Tapaswi2016}),
   our datasets do not contain any textual cues like scripts or subtitles. Also, videos are temporally continuous and are
   usually taken from a stationary view-point. This makes finding associations between objects
   across frames easy, even if datasets do not provide textual cues (\emph{e.g.} tGIF-QA
           \cite{Jang2017}). The image sets in ISVQA dataset are not akin to video frames.
   Also, unlike embodied QA \cite{Das2018}, ISVQA does not have an agent interacting with the
   environment. ISVQA algorithms can use only the few given images, which resembles real-world
   applications. Embodied QA does not require sophisticated inference of
   the correspondence between images, as the frames that an agent sees are continuous. The agent can
   reach the desired location, and answer the question using only the final frame. In contrast, ISVQA
   often needs reasoning across images and an implicit understanding of a scene.





\noindent
\textbf{VQA methods}. Most of the recent VQA methods use attention mechanisms to focus on the most
relevant image regions. For example, 
         \cite{Anderson2018} proposed a bottom-up and top-down attention mechanism for answering
         visual questions.  In addition, several methods which use co-attention (or bi-directional)
    attention over questions and images have been proposed. Such methods include
    \cite{lu2016hierarchical,Gao2019}, all of which use the information from one
    modality (text or image) to attend to the other. Somewhat different from these is the work from
    Gao \emph{et al.} \cite{Gao2019a} which proposed the multi-modality latent interaction module
    which can model the relationships between visual and language summaries in the form of latent
    vectors. 

Unlike these, \cite{Desta2018} used reasoning modules over detected objects to answer questions
about geometric relationships between objects. Similarly, Santoro \emph{et al.} \cite{Santoro2017}
proposed using Relation Networks to solve specific relational reasoning problems.  Neither of these
approaches used attention mechanisms.  In this paper, we mostly focus on attention-based mechanisms
to design the baseline models.






\section{Dataset}
\label{sec:dataset}

The main goal of our data collection effort is to aid multi-image scene understanding via question
answering. 
We use two publicly available datasets (Gibson \cite{xia2018gibson} and nuScenes \cite{Caesar2019})
as the source of images to build our datasets. 
Gibson provides navigable 3D-indoor scenes. We use the Habitat API~\cite{Savva2019} to extract
images from multiple locations and viewpoints from it. nuScences contains sets of images taken
simultaneously from multiple cameras on a car.


\subsection{Annotation Collection}
\label{sec:annot_method}

\noindent
\textbf{Indoor Scenes.}
Gibson is a collection of 3D scans of indoor
spaces, particularly houses and offices.  It provides virtualized scans of real indoor spaces like
houses and offices. 
Using the Habitat platform, we place an agent at different locations and orientations in a scene and
store the views visible to the agent. We generate a set of images by obtaining several views from
the same scene. Therefore, together, each image set can be considered to represent the scene. 

We collect two types of indoor scenes: 1.) Gibson-Building; and 2.) Gibson-Room.
Gibson-Building contains multiple images taken from the same building by placing the agent
at random locations and recording its viewpoint while Gibson-Room is
collected by obtaining several views from the same room.


For Gibson-Building, we sample image sets by placing an agent at random locations in the scene.
We show images from Gibson-Building sets to annotators and request them to ask questions about the
scene. 

We obtain question-answer annotations for a scene from several annotators using Amazon Mechanical
Turk. We let each annotator ask a question about the scene and also provide the corresponding
answer. We request that the annotators should ensure that their question can be answered using only
the scene shown and no additional knowledge should be required. 


From a pilot study, we observed that it is easier for humans to frame questions if they are shown
the full 3D view of a scene, simulating the situation of them being present in the scene. Humans are
able to frame better questions about locations of objects, and their relationships in such a
setting. For Gibson-Room, we simulate such immersion by creating a 360$^\circ$ video from a room. We
show these videos (see supplementary material for examples of how these videos are created) to the
annotators and ask them to provide questions and answers about the scenes. This process helped
annotators understand the entire scene more easily and enabled us to collect more questions
requiring across-image reasoning. Videos are not used for annotating nuScene and Gibson-Building.

Note that the ISVQA problem and datasets do not have videos. The images for Gibson-Room still came
from random views as previously described. It is possible that the image set has less coverage of
the scene than the video. Just using the image set, it might not be possible to answer the questions
collected on the video. We prune out those cases by by asking other human-annotators to verify if
the question can be answered using the provided image set. 

\noindent
\textbf{Outdoor Scenes.} 
We collect annotations for the nuScenes dataset similar to the Gibson-Building setting. We show the
annotators images from an image set. These represent a 360 degree view of a scene. We, again, ask
them to write questions and answers about the scene as before. 


\noindent
\textbf{Refining Annotations.} 
We showed all the image sets in our datasets
and the associated questions obtained from the previous step to up to three other annotators. We
asked them to provide an answer to the question based only on the image set shown. We also asked
them to say ``Not possible to answer" if the question cannot be answered. This step increases
confidence about an answer if there is a consensus among the annotators. This step has the added
benefit of ensuring that the question can be answered using the image set. 


In addition, we also asked the annotators at this stage to mark the images which are required to
answer the given question. This provides us information about which images are the most salient for
answering a question. 




\noindent
\textbf{Train and Test Splits.}
After refinement, we divided the datasets into train and test splits. The statistics of
these splits are given in table~\ref{tab:split_numbers}. For test splits, we have select 
samples for which at least two annotators agreed on the answer. We also ensured that the
train and test sets have the same set of answers.

\begin{table}
    \begin{center}
    \caption{Statistics of train and test splits of the datasets.}
    \label{tab:split_numbers}
    \resizebox{0.7\linewidth}{!}{
        \begin{tabular}{lccc}
            \toprule
            Dataset & \#Train sets & \#Test sets & \#Unique answers \\
            \midrule
            Indoor - Gibson (Room + Building) & 69,207 & 22,272 & 961 \\
            Outdoor - nuScenes & 33,973 & 15,644 & 650 \\
            \bottomrule
        \end{tabular}
    }
    \end{center}
\end{table}

\subsection{Dataset Analysis}
\label{sec:dataset_stats}



\begin{figure}
    \begin{center}
        \includegraphics[width=0.35\linewidth]{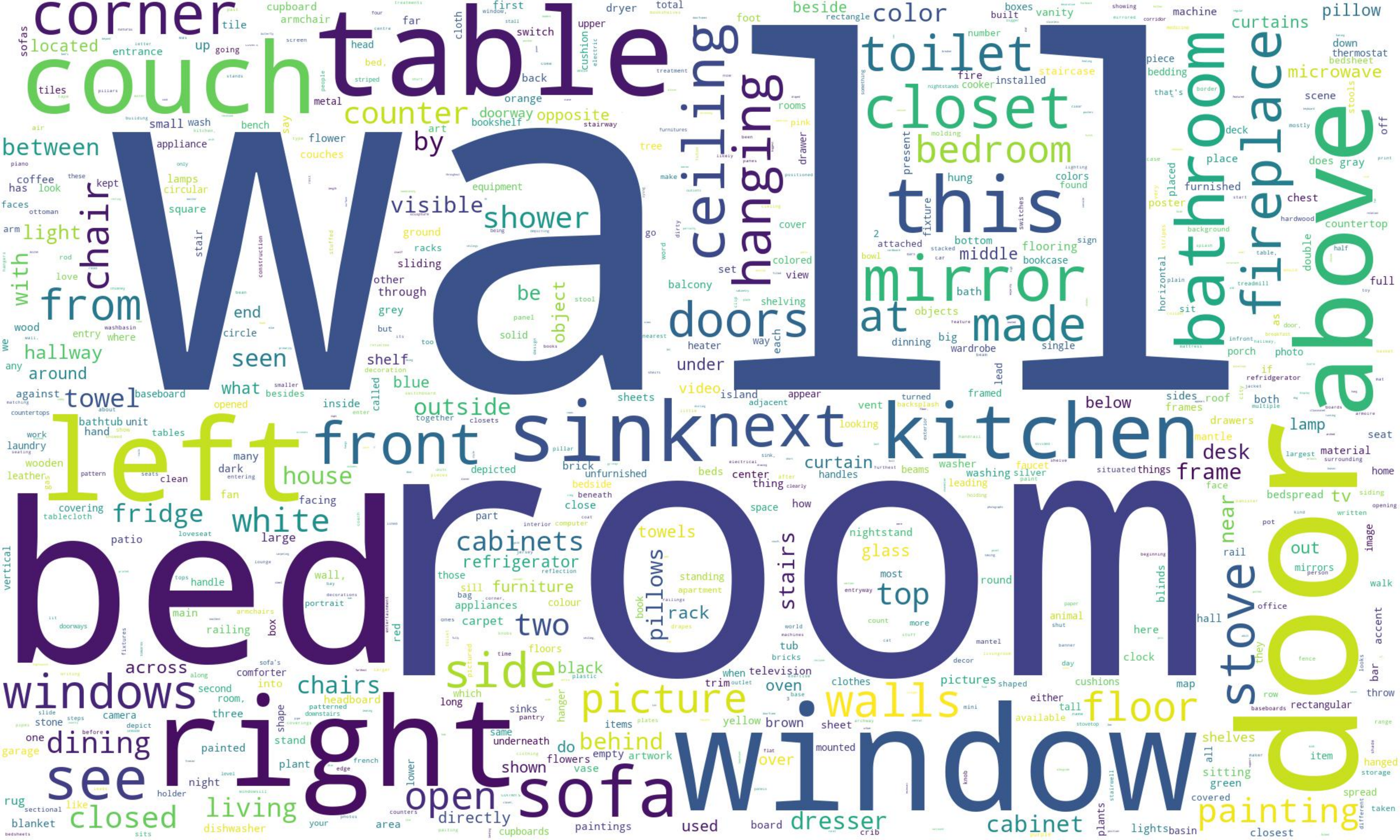}~~~~~~~~~~~~~~
        \includegraphics[width=0.35\linewidth]{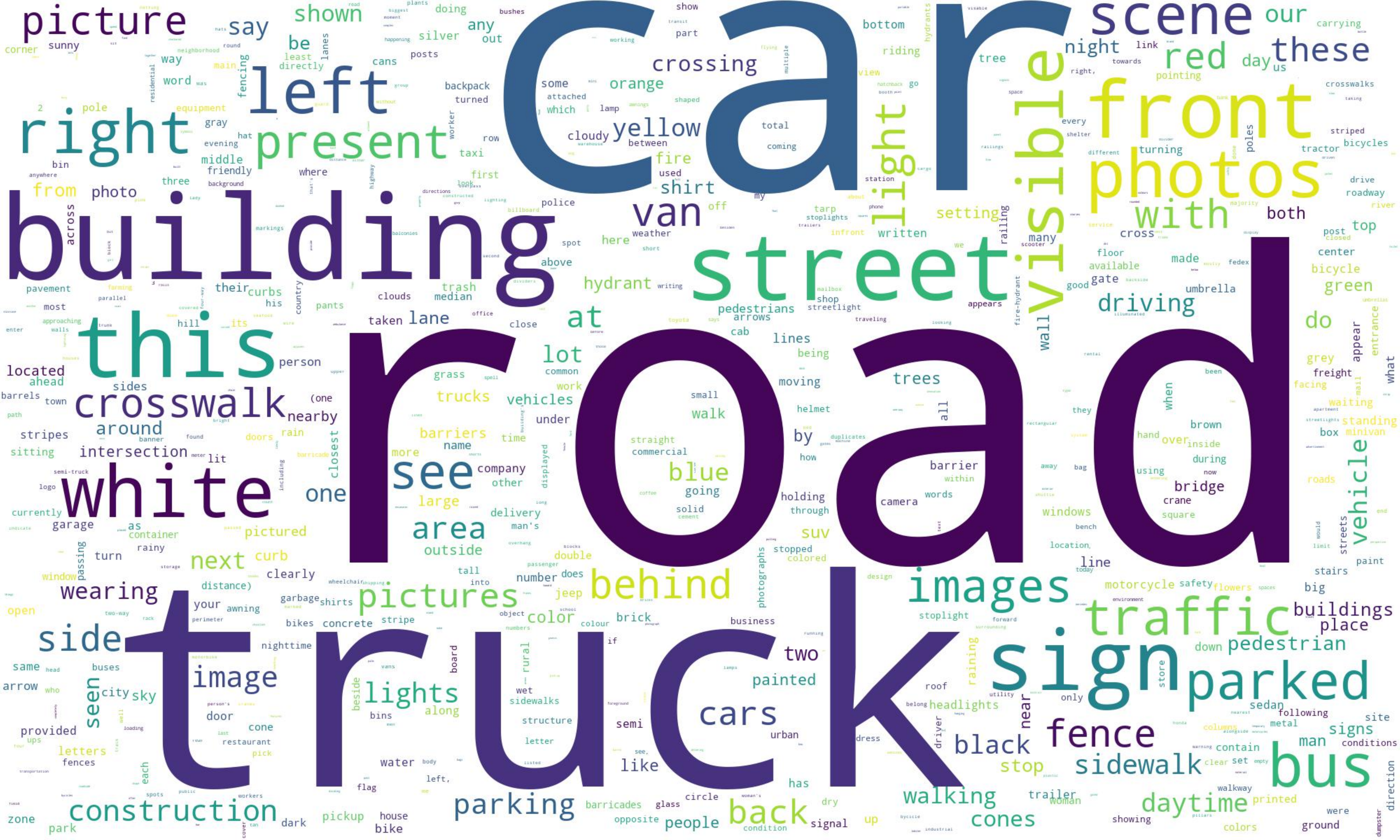}
    \end{center}
    \caption{Question wordclouds for Gibson (left) and nuScenes (right) datasets.}
    \label{fig:wordclouds}
\end{figure}

\noindent
\textbf{Question word distributions.}
The question word clouds for datasets are shown in figure~\ref{fig:wordclouds}.
We have removed the first few words from each question before plotting these. This gives us a better
picture of which objects people are interested in. 
Clearly, for outdoor scenes, people are most interested in objects commonly found on the streets
and their properties (types, colors, numbers). On the other hand, for
indoor scenes, the most frequent questions are about objects hanging on walls and kept on beds, and
the room layouts in general.

\begin{figure*}
    \centering
    \begin{subfigure}[t]{0.3\textwidth}
        \includegraphics[width=\linewidth]{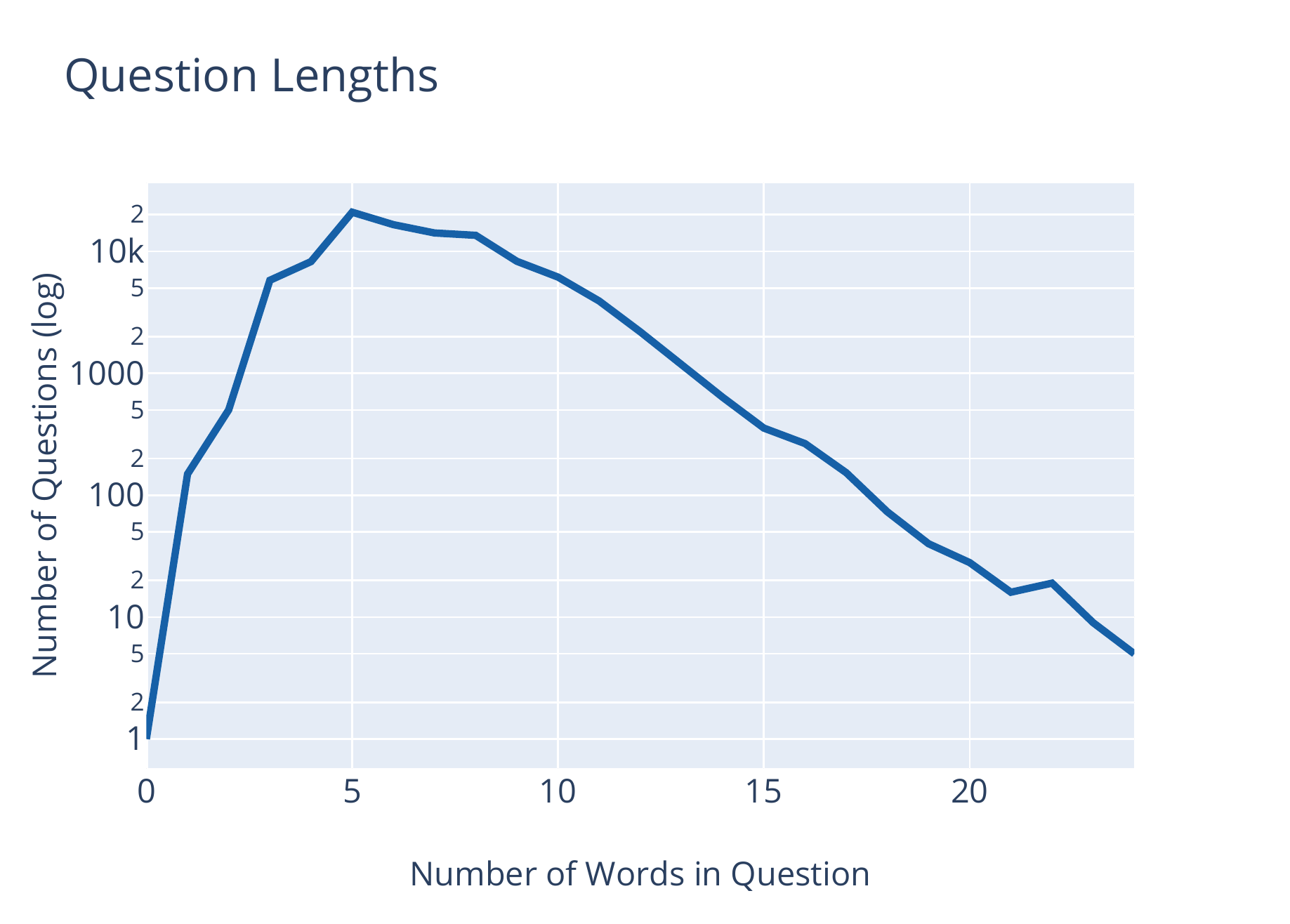}
        \caption{Question lengths}
        \label{fig:question_length}
    \end{subfigure}
    ~
    \begin{subfigure}[t]{0.3\textwidth}
        \includegraphics[width=\linewidth,height=0.85\linewidth]{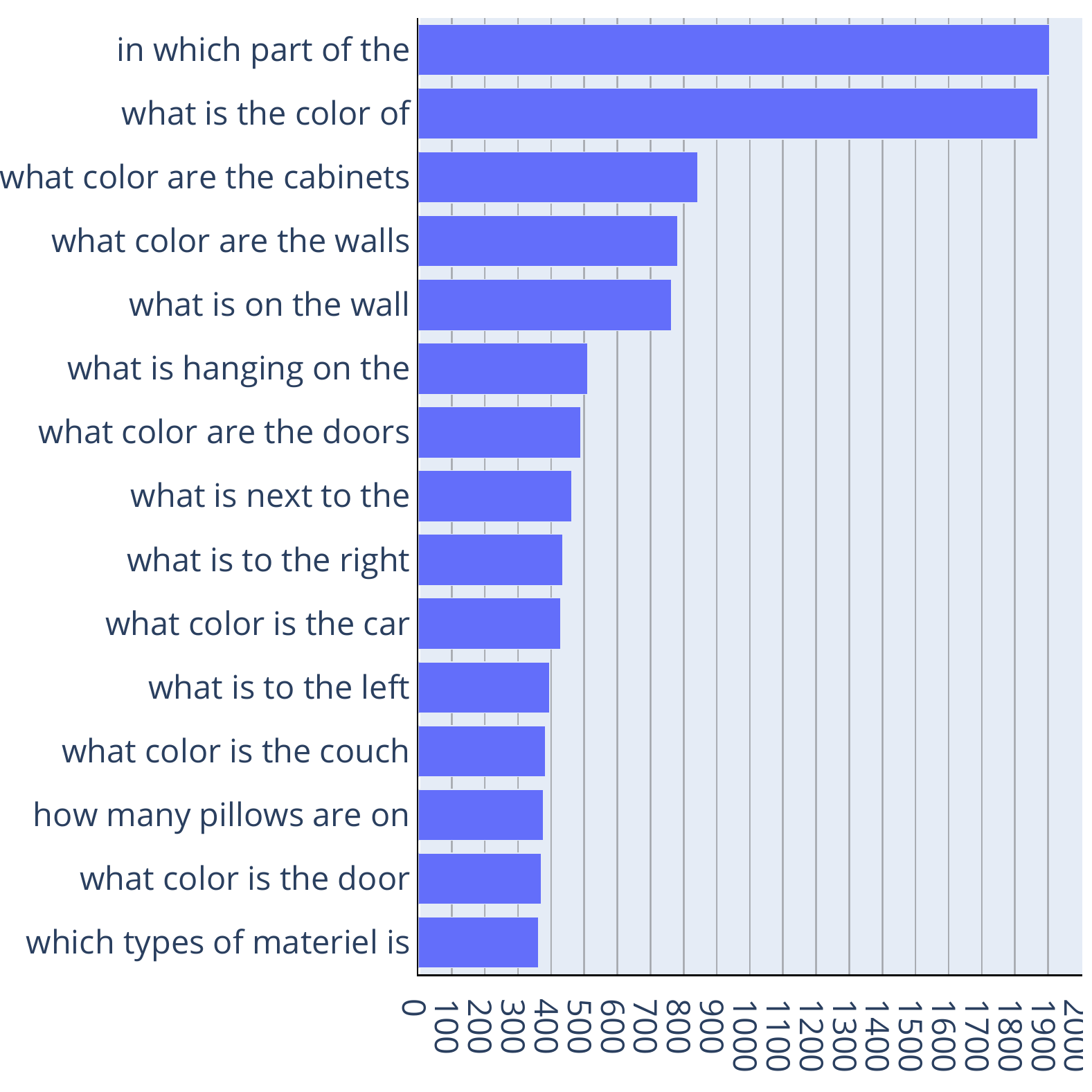}
        \caption{Frequent questions}
        \label{fig:frequent_questions}
    \end{subfigure}
    ~
    \begin{subfigure}[t]{0.3\textwidth}
        \includegraphics[trim=140 40 140 0, clip, width=0.9\linewidth]{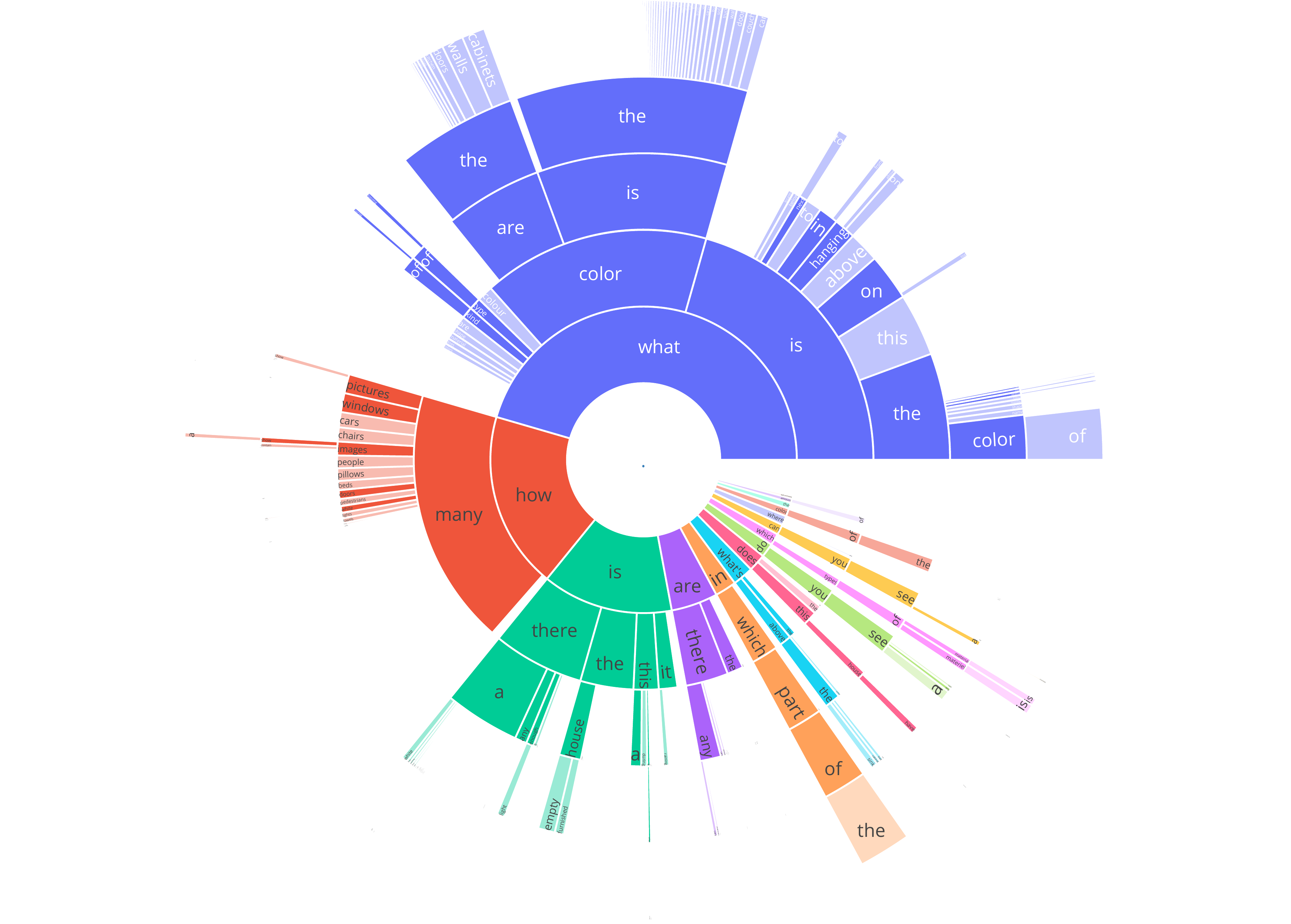}
        \caption{Types of questions}
        \label{fig:sunburst}
    \end{subfigure}
    \caption{(Left) Distribution of questions over no. of words.  (Middle) Most frequent types
        of questions in the dataset. (Right) First five words of the questions. 
    }
    \label{fig:nuscenes_plots}
\end{figure*}

\noindent
\textbf{Types of Questions.}
Figure~\ref{fig:question_length} shows the distribution of question lengths for the 
dataset. We observe that a large chunk of the questions contain between
5 and 10 words. Further, figure~\ref{fig:frequent_questions} shows the numbers of the
most frequent types of questions for the dataset. We observe that the most frequent questions are
about properties of objects, and spatial relationships between
different entities. 

To understand the types of questions in the dataset, we plot the distribution of the most frequent
first five words of the questions in the whole dataset in figure~\ref{fig:sunburst}. Note that a
large portion of the questions are about the numbers of different kinds of objects. Another major
subset of the questions are about geometric relationships between objects in a scene. A third big
part of the dataset contains questions about colors of objects in scenes. Answering questions about
the colors of things in a scene requires localization of the object of interest. Depending on the
question, this might require reasoning about the relationships between objects in different images.
Similarly, counting the number of a particular type of object requires keeping track of previously
counted objects to avoid double counting if the same object appears in different images.

 


\begin{figure}
    \centering
        \includegraphics[width=0.7\linewidth]{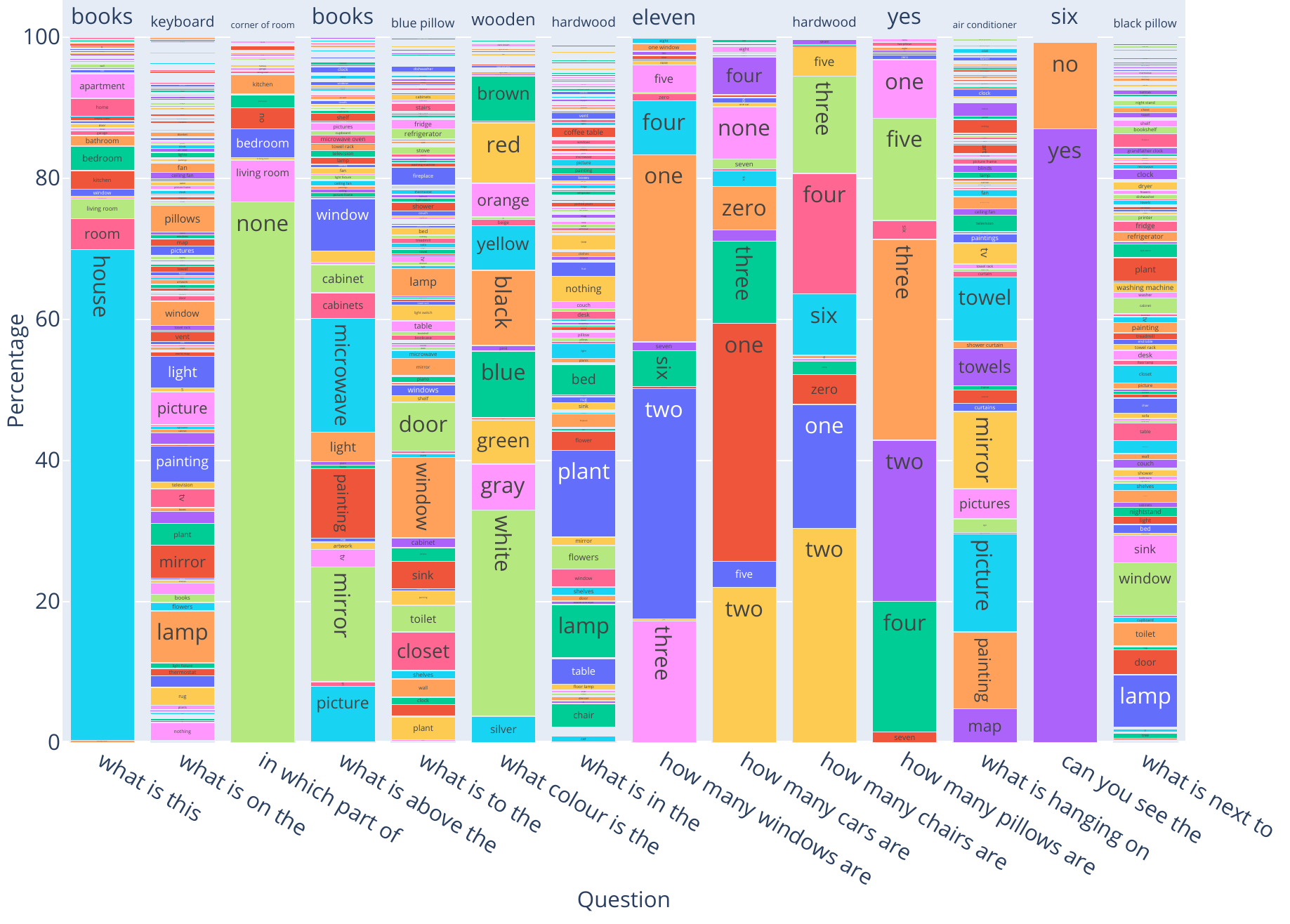}
        \caption{Answer distributions for several types of questions in the whole dataset.}
        \label{fig:answer_distributions}
\end{figure}

\noindent
\textbf{Answer Distributions.}
Figure~\ref{fig:answer_distributions} shows the distribution of answers in the dataset for
frequently occurring questions types. On one hand, due to human bias in asking questions, dominant
answers exist for a few types of questions, such as ``can you see the” (usually for an object
in the image) and ``what is this” (usually a large object). In ISVQA and other VQA
datasets humans' tendency to only ask questions about objects that they can see leads to a higher
frequency of ``yes” answers. On the other hand, most question types do not have a dominant
answer. Of particular note are the questions about relative locations and orientations of objects,
\emph{e.g.} ``What is next to", and questions about the numbers of objects
\emph{e.g.} ``How many chairs are" etc. This means that it is difficult for a
model to perform well by lazily exploiting the statistics of question types. 


\begin{figure}
    \centering
        \begin{subfigure}[t]{0.3\linewidth}
            \includegraphics[trim=0 0 40 50, clip, width=\linewidth]{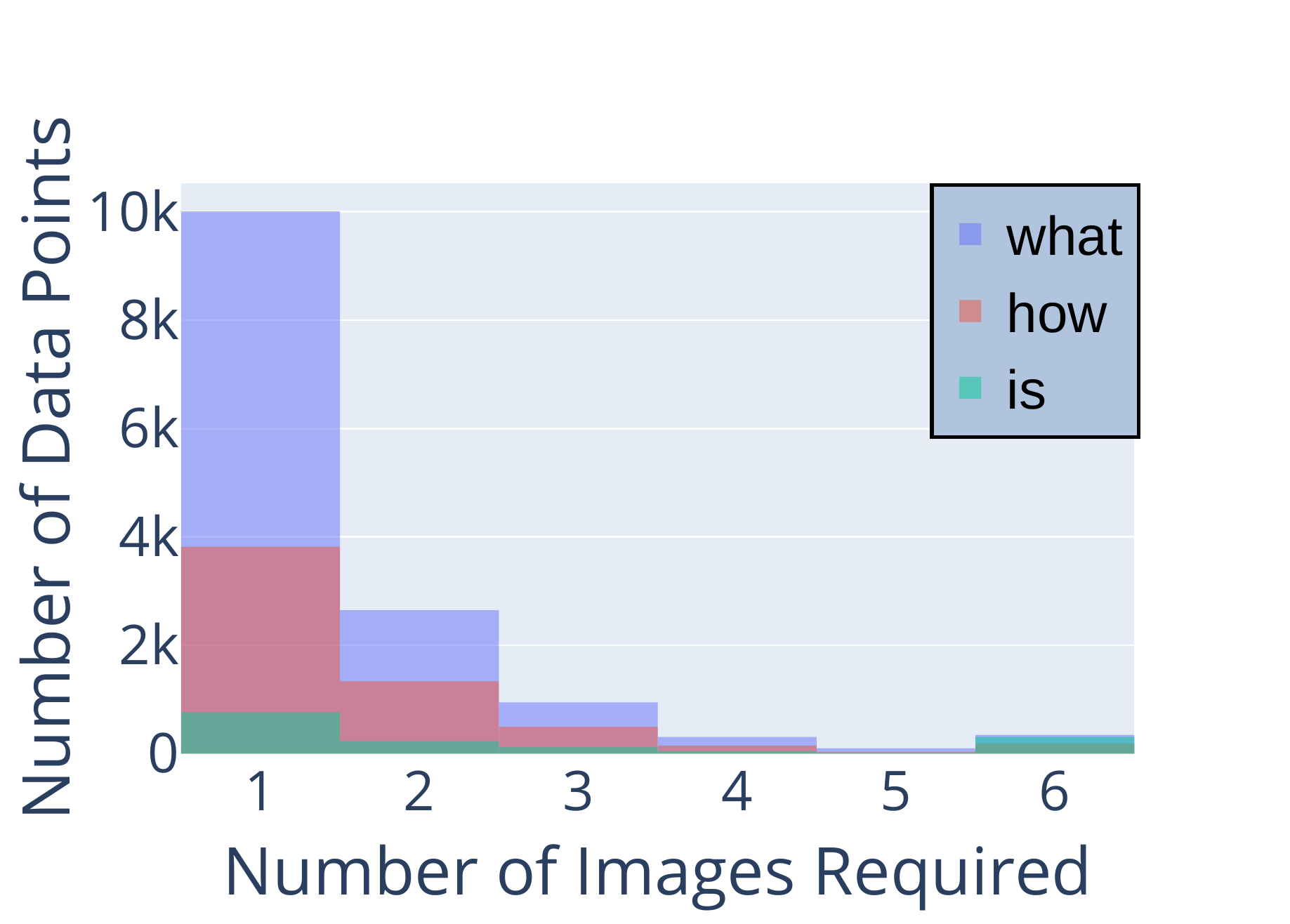}
            \caption{Gibson-Room}
        \end{subfigure}~
        \begin{subfigure}[t]{0.3\linewidth}
            \includegraphics[trim=0 0 40 50, clip, width=\linewidth]{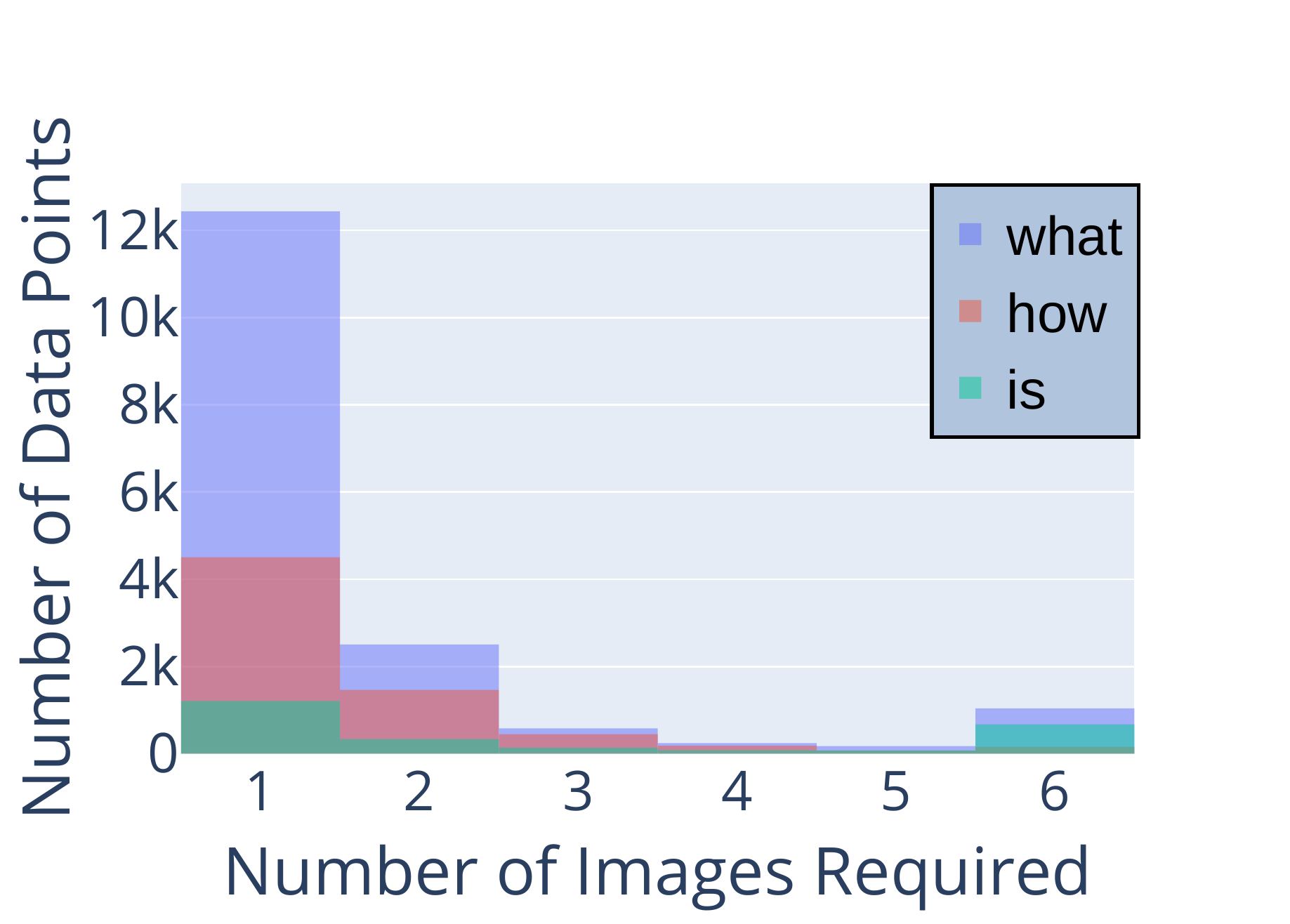}
            \caption{Gibson-Building}
        \end{subfigure}~
        \begin{subfigure}[t]{0.3\linewidth}
            \includegraphics[trim=0 0 40 50, clip, width=\linewidth]{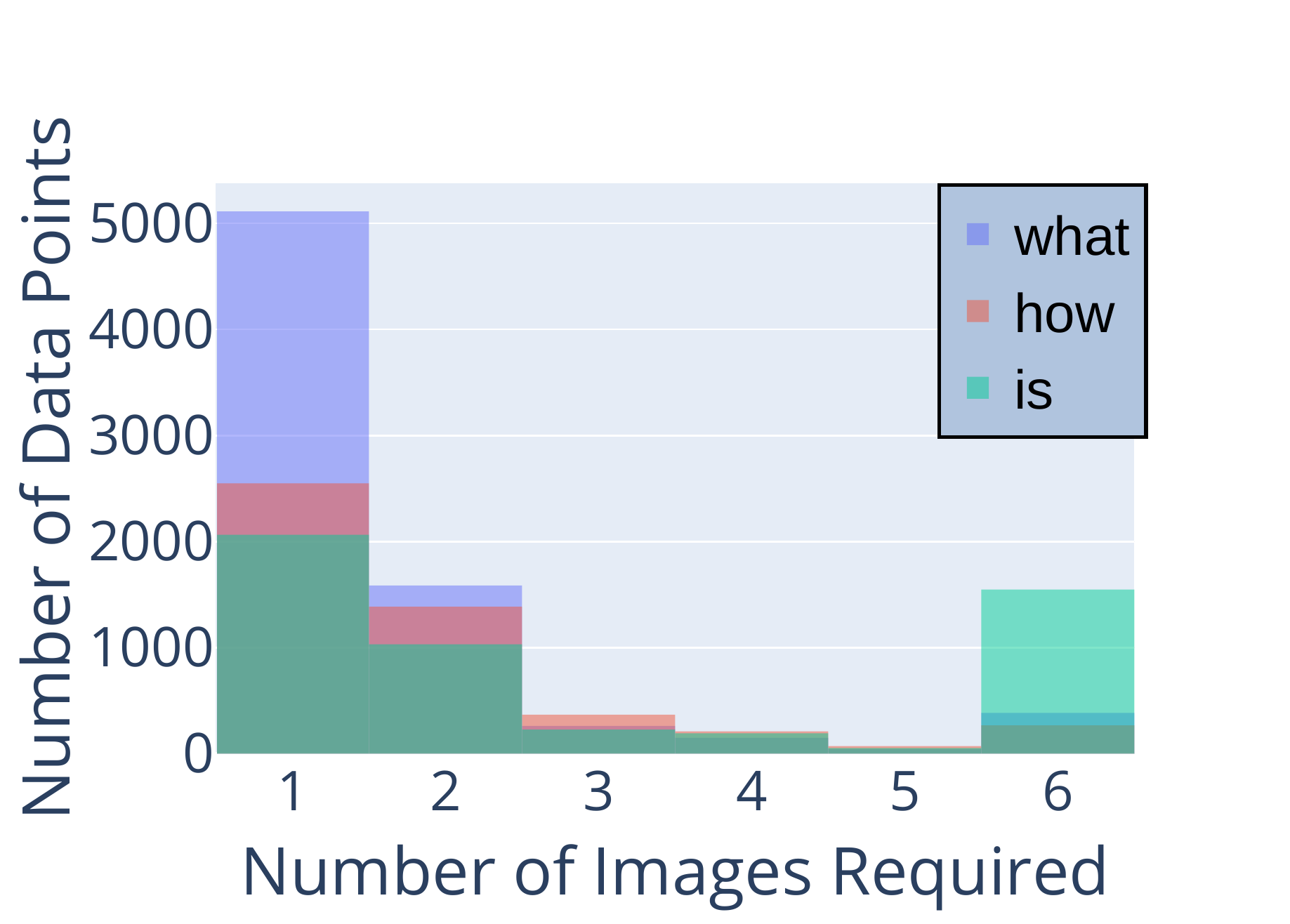}
            \caption{nuScenes}
        \end{subfigure}
        \caption{Number of images required to answer different types of questions.
        }
        \label{fig:num_required}
\end{figure}

\noindent
\textbf{Number of Images Required.}
While refining the annotations, we also collect annotations for which images are required to answer
the given question. In figure~\ref{fig:num_required}, we plot the number of images
required to answer each question for all datasets. For the plot in
figure~\ref{fig:num_required}, we only consider those image sets for which at least 2 annotators
agree about the images which are needed. 
We observe that one-third of the samples (about 7,000/21,000) in 
Gibson-Room require at least two images to answer the question. As expected, this ratio is
lower for Gibson-Building dataset. However, for all three datasets, we have a large number of
questions which require more than one image to answer. 
The large number of samples in both cases enable the study of both across-image reasoning and
image-level focusing. In particular, the latter case also involves rejecting most of the images in
the image set and focusing only on one image. In theory, such questions can potentially be answered
by using existing single-image VQA models. However, this would require the single-image VQA model to
say ``Not possible to answer" for all the irrelevant images and finding only the most relevant one.
Current VQA models do not have the ability to do this in many cases.  (see supplementary material for
examples). 



\section{ISVQA Problem Formulation and Baselines}
\label{sec:problem}


\subsection{Problem Definition}
\begin{figure}
    \centering
    \includegraphics[width=0.95\linewidth]{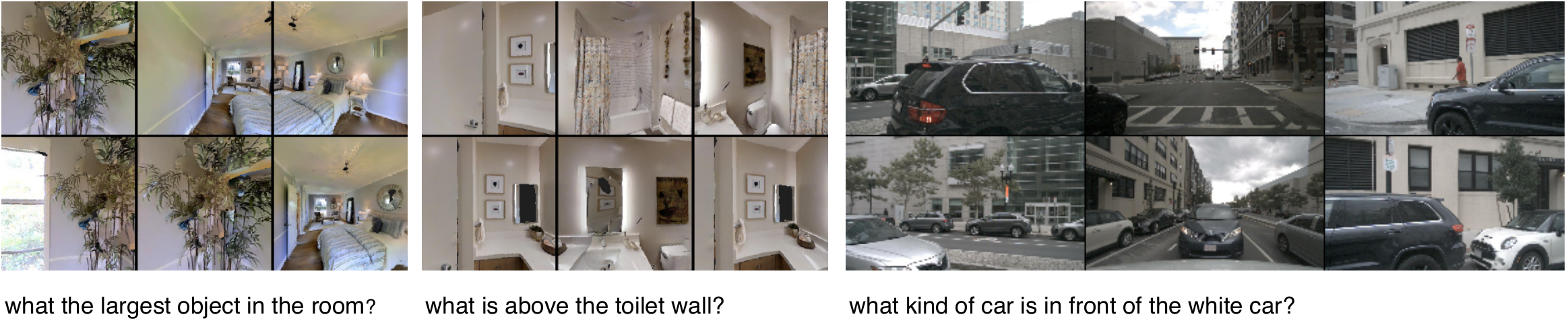}
    \caption{Some examples from our dataset which demonstrate the ISVQA problem setting. In each
    case, input is a set of images and a question.}
    \label{fig:isvqa_examples}
\end{figure}

Refer to figure~\ref{fig:isvqa_examples} for some examples of the ISVQA setting. Given a set of
images, $S = \{I_1, I_2, \dots, I_n\}$, and a natural language question, $Q = \{v_1, v_2, \dots,
v_T\}$, where $v_i$ is the $i^{th}$ word in the question, the task is to provide an answer, $a =
f(S, Q)$, which is true for the given question and image set. The function $f$ can either output a
probability distribution over a pre-defined set of possible answers, $\mathcal{A}$, or select the
best answer from several choices which are input along with the question, i.e., $a = f(S, Q, C_Q)$,
where $C_Q$ is the list of choices associated with $Q$. The former is usually called open-ended QA
and the latter is called multiple-choice QA. In this work, we mainly deal with the open-ended
setting. Another possible setting is to actually generate the answer using a text generation method
similar to image-captioning. But, most existing VQA works focus on either of the first two settings
and therefore, we also consider the open-ended setting in this work. We leave the harder problem of
generating answers to future work.

\subsection{Model Definitions}
Now, we describe some baselines for the ISVQA problem. These baselines directly adapt single image
VQA models. The first of these processes each image separately and concatenates the features
obtained from each image to predict the answer. The second baseline directly adapts VQA methods by
simply stitching the images and using single image VQA methods to predict the answer.

We also propose an approach to address the special challenges in ISVQA.
A fundamental direction to solve ISVQA problem is to enable finer-grained and across-image interactions in a VQA model, where self-attention-based transformers can fit well. 
In particular, we adapt LXMERT
\cite{tan2019lxmert}, which is designed for cross-modality learning, to both cross-modality and cross-image scenarios. 

\noindent
\textbf{Concatenate-Feature.}
Starting from a given set of $n$ images $S = \{I_1, I_2, \dots, I_n\}$, we use a region proposal
network (RPN) to extract region proposals $R_i, i = 1,2,\dots,n$ and the
corresponding RoI-pooled features (fc6).  With some abuse of notation, we denote the region features
obtained from each image as $R_i \in \mathbb{R}^{p \times d}, i = 1,2,\dots,n$, where $p$ is the
number of region features obtained from each image and $d$ is the dimension of the features. We are
also given a natural language question $Q = \{v_1, v_2, \dots, v_T\}$, where $v_i$ is the $i^{th}$
word, encoded as a one-hot vector over a fixed vocabulary $V$ of size $d_V$. For all the models, we
first obtain question token embeddings $E = \{W_w^T v_i\}_{i=1}^T$, where $W_w \in \mathbb{R}^{d_V
\times d_q}$ is a continuous word-vector embedding matrix. We obtain the question embedding feature
using an LSTM-attention module, i.e., $q = \text{AttentionPool}(\text{LSTM}(E)) \in
\mathbb{R}^{d_q}$.


\begin{figure*}
    \centering
        \includegraphics[width=0.9\linewidth]{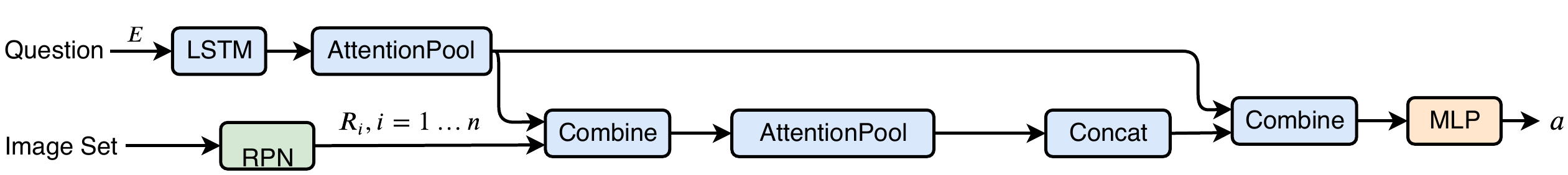}
        \caption{\textbf{Concatenate-Feature Baseline}. This method adapts a single-image VQA model
        to an image set $S = \{I_1 \dots I_n\}$. We first extract region proposals, $R_i$ from each
    image $I_i$. The model attends over the regions in each image separately using the question
embedding $q$. Pooling the region features gives a representation of an image as $x_i$. These are
concatenated and combined (element-wise multiplied) by the question embedding to give the joint
scene representation $x$. We use fully-connected layers to predict the final answer $a$.}
    \label{fig:models}
\end{figure*}

Figure~\ref{fig:models} shows an outline of the model.  For each image, $I_i$, we obtain the image
embedding, $x_i$ by attending over the corresponding region features $R_i$ using the question
embedding $q$.
\begin{equation}
    \label{eq:image_feat}
    x_i = \text{AttentionPool}(\text{Combine}(R_i, q))
\end{equation}
where, we use element-wise multiplication (after projecting to suitable dimensions) as the
$\text{Combine}$ layer and $\text{AttentionPool}$ is a combination of an $\text{Attention}$ module
over the region features which is calculated through a softmax operation and a $\text{Pool}$
operation. The region features are multiplied by the attention and added to obtain the pooled image
representation. For a single image, this model is an adaptation of the recent Pythia model
\cite{singh2019TowardsVM} without its OCR functionality. We concatenate the image features $x_i$ and
element-wise multiply by the question embedding to obtain the joint embedding
\begin{equation}
    x = \text{Combine}(\text{Concat}(x_1, x_2, \dots , x_n), q)
\end{equation}
where the $\text{Combine}$ layer is again an element-wise multiplication. This is passed through a
small MLP to obtain the distribution over answers, $P_A = \text{MLP}(x)$.

\noindent
\textbf{Stitched Image.}
Our next baseline is also an adaptation of existing single-image VQA methods. We start by
stitching all the images in an image set into a mosaic, similar to the ones shown in figure~\ref{fig:isvqa_examples}. Note that the ISVQA setting does not require the images in an image set to
follow an order. Therefore, the stitched image obtained need not be panoramic. We train the recent
Pythia \cite{singh2019TowardsVM} model on the stitched images and report performance in table~\ref{tab:results}.

\noindent
\textbf{Video VQA.}
To highlight the differences between Video VQA and ISVQA, we adapt the recent state-of-the-art
method HME-VideoQA \cite{hme2019}. This model consists of heterogeneous memory module which can
potentially learn global context information. We consider images in the image set as frames of a
video. Note that, the images in an image set in ISVQA do not necessarily constitute the frames of a
video. Therefore, it is reasonable to expect such Video VQA methods to not provide any advantages
over our baselines. 

Using these baselines, we show that ISVQA is not a trivial extension of VQA. Solving ISVQA requires development of specialized methods. 

\noindent
\textbf{Transformer-based Method.}
We utilize the power of transformers and adapt the LXMERT model~\cite{tan2019lxmert} to both
cross-modality and cross-image scenarios. The transformer can summarize the relevant information
within an image set and also model the across-image finer-grained dependencies. Here, we briefly
described the original LXMERT model and then describe our modifications.

LXMERT learns cross-modality representations between regions in an image and sentences. It first
uses separate visual and language encoders to obtain visual and semantic embeddings. The visual
encoder consists of several self-attention sub-layers which help in encoding the relationships
between objects. Similarly, the language encoder consists of multiple self-attention sub-layers and
feed-forward sub-layers which provide a semantic embedding for the sentence or question. The visual
and semantic embeddings are then used to attend to each other via cross-attention sub-layers. This
helps the LXMERT model learn final visual and language embeddings which can tightly couple the
information from visual and semantic domains. These coupled embeddings can be seen as the joint
representations of the image and sentence and are used for inference.

Instead of using features from only a single image as input to the object-relationship encoder, we
propose to use the region features from each image in our image-set. As described above, we start by
extracting $p$ region proposals and the corresponding features from each of the $n$ images in the
image set. We pass the $p \times n$ region features as inputs to the object-relationship encoder in
LXMERT. We note that this enables the our model to encode relationships between objects across
different images. 

Let us denote the image features as $R = [R_1; R_2; \dots; R_n] \in \mathbb{R}^{pn \times d}$, where $R_i$ are the region features obtained from $I_i$. We also have the corresponding
position encodings of each region in the images $P = [P_1; P_2; \dots; P_n] \in \mathbb{R}^{pn
\times 4}$, where $P_i$ contains the bounding box co-ordinates of the regions in $I_i$. We
combine the region features and position encodings to obtain position-aware embeddings, $S \in
\mathbb{R}^{pn \times d'}$, where $S = \text{LayerNorm}(\text{FC}(R)) +
\text{LayerNorm}(\text{FC}(P))$. Within- and across-image object relationships are encoded by applying $N_R$ layers of the object relationship encoder. The $l$-th layer can be represented as
\begin{equation}
    \label{eq:object_relationship}
    x_l = \text{FC}(\text{FC}(\text{SelfAttention}(x_{l-1})))
\end{equation}
where, $x_0 = S$, and $X (= x_{N_R})$ is the final visual embedding of the object-relationship
encoder.

Similarly, given the word embeddings of the question, $E$, and the index embeddings of each word in
the question, $E' = \{\text{IdxEmbed}(1), \dots, \text{IdxEmbed}(T)\}$, the
index-aware word embedding of the $i$-th word is obtained as $H_i =\\ \text{LayerNorm}(E_i
+ E_i^{'})$. Note that the index embedding, $\text{IdxEmbed}$, is just a
projection of the position of the word to a vector using fully-connected layers. We apply a
similar operation as equation \ref{eq:object_relationship} $N_L$ times to the word embeddings $H = [H_1;
H_2; \dots; H_T] \in \mathbb{R}^{T \times d_q}$ to give the question embedding, $L$.

Finally, LXMERT consists of $N_X$ cross-modality encoders stacked one-after-another. Each
encoder consists of two operations: 1.) language to vision cross attention,
$X = \text{FC}(\text{SelfAttention}(\text{CrossAttention}_{LV}(X, L)))$; and 2.) vision to
language cross attention, $L =
\text{FC}(\text{SelfAttention}(\text{CrossAttention}_{VL}(L,\allowbreak X)))$.
The final output of the $N_X$ encoders are used to predict the answer.

\noindent
\textbf{Evaluating Biases in the Datasets.}
We also evaluate the following prior-based baselines to reveal and understand the biases present in
the datasets.

\noindent
\underline{Na\"ive Baseline}. 
The model always predicts the most frequent answer from the training set. For nuScenes, it always
predicts ``yes", while for Gibson it predicts ``white". Ideally, this should set a minimum
performance bar. 

\noindent
\underline{Hasty-Student Baseline}. In this baseline, a model simply finds the most frequent answer
for each type of question. In this case, we define a ``question type" as the first two words of a
question. For example, a hasty-student might always answer ``one" for all ``How many" questions.
This is similar to the hasty-student baseline used in \cite{Liang2018} (MovieQA).  

\noindent
\underline{Question-Only Baseline}. In this model, we ignore the visual information and only use
question text to train a model. Our implementation takes as input only the question embedding, $q$
which is passed through several fully-connected layers to predict the answer distribution. This
baseline is meant to reveal the language-bias present in the dataset.

            

\section{Experiments}
\label{sec:experiments}


\subsection{Human Performance}
An ideal image-set question answering system should be able to reach at least the accuracy achieved
by humans. We evaluate the human performance using the annotations with the standard VQA-accuracy
metric described below. For the outdoor scenes dataset, humans obtain a VQA-accuracy of 91.88\% and
for the indoor scenes they obtain 88.80\%. Comparing this with table \ref{tab:results}
shows that ISVQA is extremely challenging and requires specialized methods. The reason for
the human performance being lower than 100\% is that, in many cases an annotator has given an answer
which is not exactly similar to the other two but is still semantically similar. For example, the
majority answer might have been ``black and white" but the third annotator answered ``white and
black". 


\subsection{Implementation Details}
\label{sec:implementation}
We start by using Faster R-CNN in Detectron to extract the region proposals and features $R_i$
for each image. Each region feature is 2048-D and we use the top 100 region proposals from each
image. To obtain the word-vector embeddings we use 300-D GloVe \cite{pennington2014glove} vectors.
The joint visual-question embedding, $x$ is taken to be 5000-D. For evaluation, we use the VQA-Accuracy 
metric \cite{Antol2015}. A predicted answer is given a score of one if it matches at least two out
of the three annotations. If it matches only one annotations, it is given a score of 0.5.
All of our VQA models are implemented in the Pythia framework \cite{singh2018pythia} and are trained
on two NVIDIA V100 GPUs for 22,000 iterations with a batch size of 32. The initial learning rate is
warmed up to $0.01$ in the first 1,000 iterations. The learning rate is dropped by a factor of 10 at
iterations 12,000 and 18,000. For the HME-VideoQA model, we use the implementation provided by the
authors. We train the model for 22,000 iterations with a batch size of 32 and a starting learning
rate of 0.001. For the transformer-based model, we use $N_L = 9, N_R = 5, N_X = 5$. We use a
batch-size of 32, learning rate of 0.00005, and we train the model for 20 epochs. All feature
dimensions are kept the same as LXMERT.

\subsection{Results}
\label{sec:results}
We report the VQA-accuracy for all methods in table~\ref{tab:results}. The accuracy achieved by
both of the VQA-based baselines is only around $50-54\%$ and the Video VQA model achieves only
$39.88\%$ for the indoor dataset and $52.14\%$ for the outdoor dataset. This highlights the need for
advanced models for ISVQA. 

\noindent
\textbf{Comparison between Baselines.}
Table~\ref{tab:results} shows that the na\"ive baseline reaches a VQA-Accuracy of only 8.6\% for the
indoor scenes dataset compared to 47.57\% given by the Concatenate-Feature baseline and 50.53\%
given by the Stitched-Image baseline model. This shows that single-image VQA methods are not enough
to overcome the challenges presented by ISVQA. On the other hand, our proposed transformer-based
model performs the best for both indoor and outdoor scenes out-performing other models by over
10\%. 

\noindent
\textbf{Language Biases.}
Recent works (\emph{e.g.} \cite{Agrawal2016}) show that high performance in VQA could be achieved
using only the language components. Deep networks can easily exploit biases in the datasets to find
short-cuts for answering questions. We observe that most VQA-based
baselines perform much better than the question-only baseline. This shows that the ISVQA datasets
are less biased compared to many existing VQA datasets \cite{Agrawal2016} and validates the
utility of developing ISVQA models that can utilize both the visual and language components
simultaneously. 



\begin{table}
    \centering
        \caption{Results for both indoor and outdoor datasets.} 
        \label{tab:results}
        \resizebox{0.65\linewidth}{!}{
        \begin{tabular}{ccc|c}
            \toprule
            & \textbf{Method} & \multicolumn{2}{c}{\textbf{VQA-Accuracy} (\%)} \\
            & & Gibson & nuScenes \\
            \midrule
            & Na\"{i}ve & 8.61 & 22.46 \\
            \textbf{Prior-Based Baselines} & Hasty-Student & 27.22 & 41.65 \\
            & Question-Only & 40.26 & 46.06 \\
            \midrule
            & Video-VQA & 39.88 & 52.14 \\
            \textbf{Approaches} & Concatenate-Feature & 47.57 & 53.66 \\
            & Stitched-Image & 50.53 & 54.32 \\
            & Transformer-Based & 61.58 & 64.91 \\
            \midrule
            \textbf{Human Performance} &  & 88.80 & 91.88 \\
            \bottomrule
        \end{tabular}
    }
\end{table}

%
%
%
%
%
%

\begin{figure}
    \centering
    \includegraphics[trim=20 15 15 70, clip, width=0.75\linewidth]{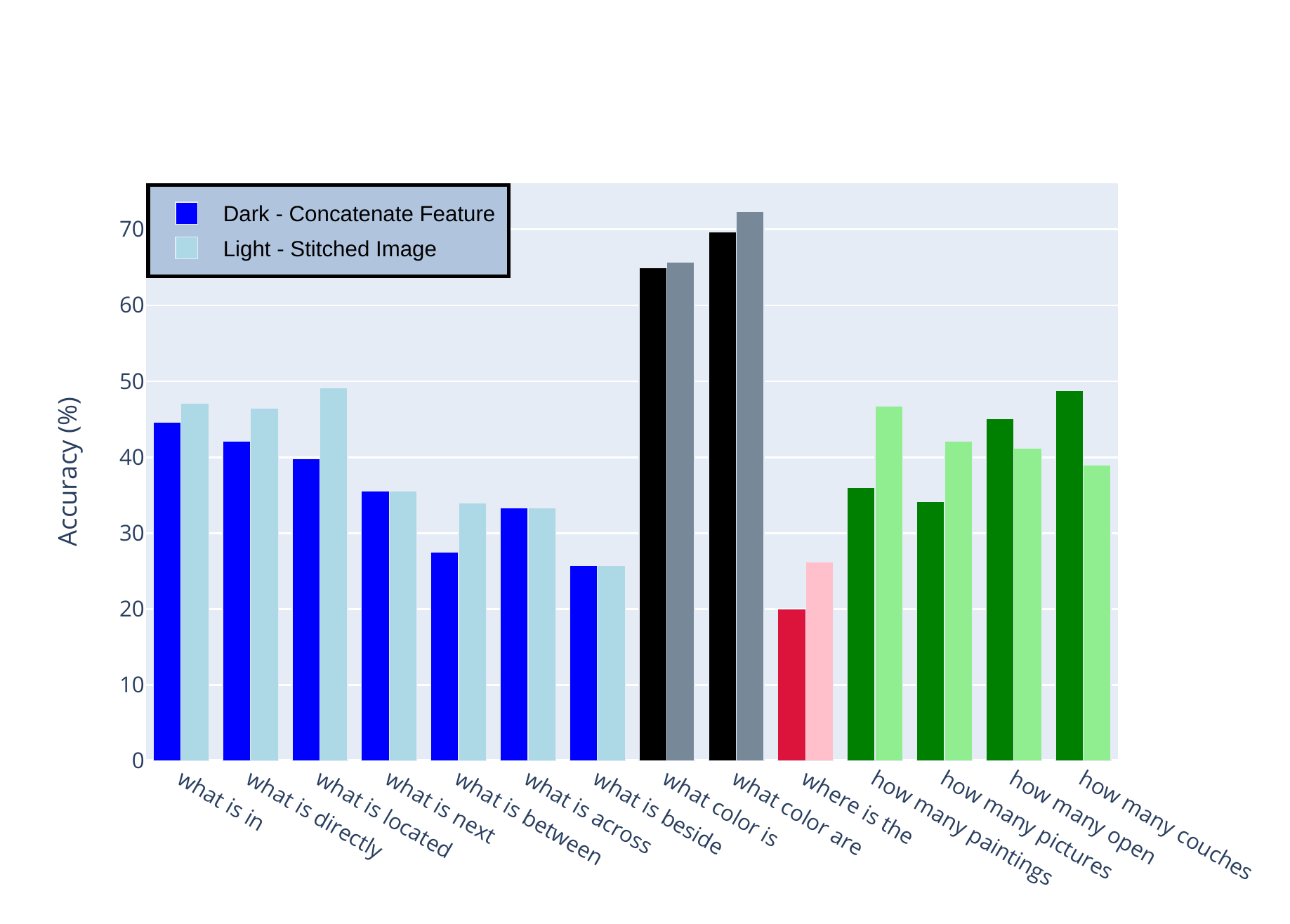}
    \caption{Performance of the two VQA-based baselines for different types of questions for the
        combined Gibson test set. Dark
        colors represent the performance for Concatenate-Feature baseline and light colors for Stitched Image
        baseline. \textcolor{blue}{\textbf{Blue}} is used for geometric relationship questions,
        \textcolor{ForestGreen}{\textbf{green}} for counting questions,
        \textcolor{red}{\textbf{red}} for location, and \textbf{black} for color questions. We
        notice that the VQA-based baselines are able to answer simple questions like those about
        colors of single objects very well. However, questions involving spatial reasoning between
    objects in one image or across images are extremely challenging for such methods.}
    \label{fig:perf_by_ques_type}
\end{figure}

\noindent
\textbf{Performance by Question Type.}
Figure~\ref{fig:perf_by_ques_type} shows the accuracy bar-chart of our single-image VQA-based
baselines for various types of questions. Using this chart, we have the following observations and
hypotheses:



\noindent
\textbf{Single-image VQA baselines can predict single-object attributes.}
Both baseline models can answer questions about colors of single objects well (black and gray bars).
This is expected because no cross-image dependency is needed. 

\noindent
\textbf{General cases may need cross-image inference. }
A large portion of questions involve multiple objects, which may appear in different images. The two VQA baselines using simple attention do not perform well on such questions. 
The across-image transformer-based approach performs much better.

\noindent
\textbf{Stitched-Image captures cross-image dependency better.}
The Stitched-Image baseline allows direct pooling from regions in all images, 
which may capture across-image dependency better. 
It also outperforms the
Concatenate-Feature baseline for most question types, except for the counting questions.
The Stitched-Image cannot avoid double counting. 
The transformer-based approach has all the advantages of the Stitched-Image and can do more sophisticated inference. 


\section{Conclusion and Discussion}
\label{sec:conclusion}

We proposed the new task of image-set visual question answering (ISVQA). 
This task can lead to new research challenges, such as language-guided cross-image attentions and reasoning. 
To establish the ISVQA problem and enable its research, we introduced two ISVQA datasets for indoor
and outdoor scenes.  Large-scale annotations were collected for questions and answers with novel
ways to present the scene to the annotators.  We performed bias analysis of the datasets to set up
performance lower bounds.  We also extended a single-image VQA method to two simple attention-based
baseline models and showed the performance of state-of-the-art Video VQA model.  Their limited
performance reflects the unique challenges of ISVQA, which cannot be solved trivially by the
capabilities of existing models. Approaches for solving the ISVQA problem may need to pass
information across images in a sophisticated way, understand the scene behind the image set, and
attend the relevant images. Another potential direction could be to create explicit maps of the
scenes. However, depending on the complexity of the scene, different techniques might be required to
explicitly construct a coherent map. Where such maps can be obtained accurately,
reconstruction-based ISVQA solutions can be more accurate than the baselines. Meanwhile,
humans do not have to do exact scene reconstruction to answer questions. So, in this
paper, we have focused on methods that can model across-image dependencies implicitly.

{\footnotesize
    \bibliographystyle{splncs04}
    \bibliography{vqa_refs,other_refs}
}

\end{document}